\documentclass[conference]{IEEEtran}
\IEEEoverridecommandlockouts
\pdfoutput=1
\usepackage{amsmath,amssymb,amsfonts}
\usepackage{physics}
\usepackage{pdfpages}
\usepackage{algorithmic}
\usepackage{graphicx}
\usepackage{multirow}
\usepackage{adjustbox}
\usepackage{wrapfig}
\usepackage{caption}
\usepackage{subcaption}
\usepackage{booktabs}
\usepackage[maxbibnames=99, sorting=none]{biblatex}

\graphicspath{ {./images/} }
\usepackage{textcomp}
\usepackage{xcolor}

\DeclareRobustCommand*{\IEEEauthorrefmark}[1]{%
  \raisebox{0pt}[0pt][0pt]{\textsuperscript{\footnotesize #1}}%
}

\def\BibTeX{{\rm B\kern-.05em{\sc i\kern-.025em b}\kern-.08em
    T\kern-.1667em\lower.7ex\hbox{E}\kern-.125emX}}
\begin{document}

\title{Multimodal Adaptive Fusion of Face and Gait Features using Keyless attention based Deep Neural Networks for Human Identification\\
{\footnotesize}
\thanks{}
}
\author{\IEEEauthorblockN{Ashwin Prakash\IEEEauthorrefmark{1},
Thejaswin S\IEEEauthorrefmark{1} and 
Athira Nambiar\IEEEauthorrefmark{1}\textsuperscript{*},
Alexandre Bernardino\IEEEauthorrefmark{2}}
\IEEEauthorblockA{\IEEEauthorrefmark{1}Department of Computational Intelligence, SRM Institute of Science and Technology, Chennai, India.\\
\{ap4471, ts6959, athiram\}@srmist.edu.in}
\IEEEauthorblockA{\IEEEauthorrefmark{2}Department of Electrical and Computer Engineering, Instituto Superior Técnico, Univ. of Lisbon, Portugal.\\
alex@isr.tecnico.ulisboa.pt }}

\maketitle

\begin{abstract}
Biometrics plays a  significant role in vision-based surveillance applications. Soft biometrics such as gait is widely used with face in surveillance tasks like person recognition and re-identification. Nevertheless,  in practical scenarios, classical fusion techniques respond poorly to changes in individual users and in the external environment. To this end, we propose a novel adaptive multi-biometric fusion strategy for the dynamic incorporation of gait and face biometric cues by leveraging keyless attention deep neural networks. Various external factors such as viewpoint and distance to the camera, are investigated in this study. Extensive experiments have shown superior performance of the proposed model compared with the state-of-the-art model.

\end{abstract}

\vspace{-0.3cm}
\section{Introduction}

Human biometrics refers to the unique intrinsic physical or behavioural traits that allow distinguishing between different individuals, e.g.,  face, fingerprint, hand geometry, iris, and gait. The use of biometrics helps in various surveillance applications such as access control, human recognition, and re-identification. Single biometric modalities are often affected by practical challenges such as noisy data, lack of distinctiveness, intra/ inter-class variability, error rate, and spoof attacks. A common method to overcome this issue is to combine multiple biometric modalities, known as multimodal biometric fusion.

A critical constraint that any biometric system confronts is the variation in the environment owing to external conditions. This includes user-induced variability, i.e., inherent distinctiveness, pose, distance, and expression, or environment-induced variability, i.e., lighting condition, background noise, and weather conditions~\cite{NAP12720}. These constraints have not been adequately addressed in the literature on multimodal fusion. For instance, most of the existing works are based on static fusion strategies, wherein the fusion rules are fixed for certain external conditions such as pose/ lighting/ distance or based on manual computations. As a result, when the environment changes, the biometric system performs sub-optimally. To overcome this issue, a novel context-aware adaptive multibiometric fusion strategy, which can dynamically adapt the fusion rules to external conditions, is proposed in this paper. In particular, the adaptive fusion of gait and face at different viewpoints was investigated using an attention-based deep learning technique.


Face is one of the predominant biometric traits commonly employed in human recognition. Similarly, gait is an important soft biometric commonly used in surveillance applications, because it is unobtrusive and perceivable from a distance~\cite{singh2021survey}. While fusing gait and face, the most influential factors may be the view angle and distance from the subject to the camera. Notably, gait can be clearly captured in the lateral view, whereas the face can be well captured in the frontal view. Based on this rationale, a novel context-aware adaptive fusion mechanism was designed to assign weights to gait and face biometric cues based on the context. The key notion of the proposed model is that when the person is in far/ lateral view, gait features should be gaining more priority than the less visible facial cues, whereas when the person is in near/ frontal view, the face should be getting more importance than the partially occluded gait features. 

To facilitate the aforesaid context-aware adaptive fusion strategy, a keyless attention-based deep learning fusion is leveraged in the multimodal biometric fusion framework. As mentioned in~\cite{long2018multimodal}, keyless attention is a sophisticated and effective technique for better accounting for the sequential character of data without the need for supplementary input, thereby excelling in identifying relationships across modalities. Extensive experiments are conducted via individual biometric-based identification, na\"ive bilinear pooling~\cite{lin2015bilinear} based multimodal fusion and keyless attention-based adaptive fusion mechanism. Results clearly highlight the superior performance of the proposed model.



The remainder of this paper is organized as follows. Related works on face and gait-based human recognition are detailed in Section 2. Section 3 presents the framework of the proposed context-aware adaptive multibiometric fusion method. The experiments and results are presented in Sections 4 and 5, respectively. Finally, conclusions and future directions are presented in Section 6.

\section{Related Work}

One of the earliest face recognition systems was discovered in~\cite{mm1966} using the manual marking of various facial landmarks. Recognition of faces in images with objects has gained popularity with~\cite{turk1991eigenfaces}, which introduced the eigenface method. Since then, various other similar techniques, e.g., Linear Discriminant Analysis, to produce
Fisherfaces,  Gabor, LBP, and PCANet were reported in ~\cite{wang2021deep}. Recently, deep learning-based techniques have also gained popularity, e.g. 
DeepFaces, Facenet, and Blazeface approach human-level performance under unconstrained conditions~\cite{wang2021deep} (DeepFace: 97.35\% vs. Human: 97.53\%).


Classical gait-based identification approaches use either model-based or appearance-based approaches\cite{singh2021survey}. The former detects joints/body parts using 2D cameras or depth cameras. For example,~\cite{cunado2003automatic} applied Hough transform to detect legs in each frame, whereas~\cite{wang2004fusion} leveraged Procrustes shape analysis to calculate joint angles of body parts. Gait recognition/re-identification using a Kinect camera has also been proposed in some works ~\cite{nambiar2017towards}. 
In contrast to model-based approaches, appearance-based approaches use richer information, such as silhouettes of the human body in gait frames, to recognise gaits, e.g., gait energy image (GEI)~\cite{han2005individual} and GEI-based local multi-scale feature descriptors~\cite{lishani2019human}. Recent deep learning approaches presented advanced techniques, e.g., view-invariant gait recognition using a convolutional neural network GEINet~\cite{shiraga2016geinet}, a comprehensive model with both LSTM and residual attention components for cross-view gait recognition~\cite{li2019attentive}.


On the fusion of gait and face for human identification, one of the early works~\cite{kale2004fusion} proposed a fusion strategy by combining the results of gait and face recognition algorithms based on sequential importance sampling. A probabilistic combination of facial and gait cues was studied in~\cite{shakhnarovich2002probabilistic}. Yet another work on the adaptive fusion of gait and face is~\cite{geng2008adaptive} via score-level fusion. All the aforementioned studies leverage either classical machine learning techniques using handcrafted features, static fusion rules, or manual computations. On the contrary, in this work, we present a deep learning technique based on a keyless attention-based adaptive fusion mechanism for human identification, one of its first kind to the best of our knowledge. 





\section{Multimodal Adaptive Fusion Methodology }
\label{METHOD}
\begin{figure*}[htbp]
\begin{center}
\includegraphics[width=14cm, height=7cm]{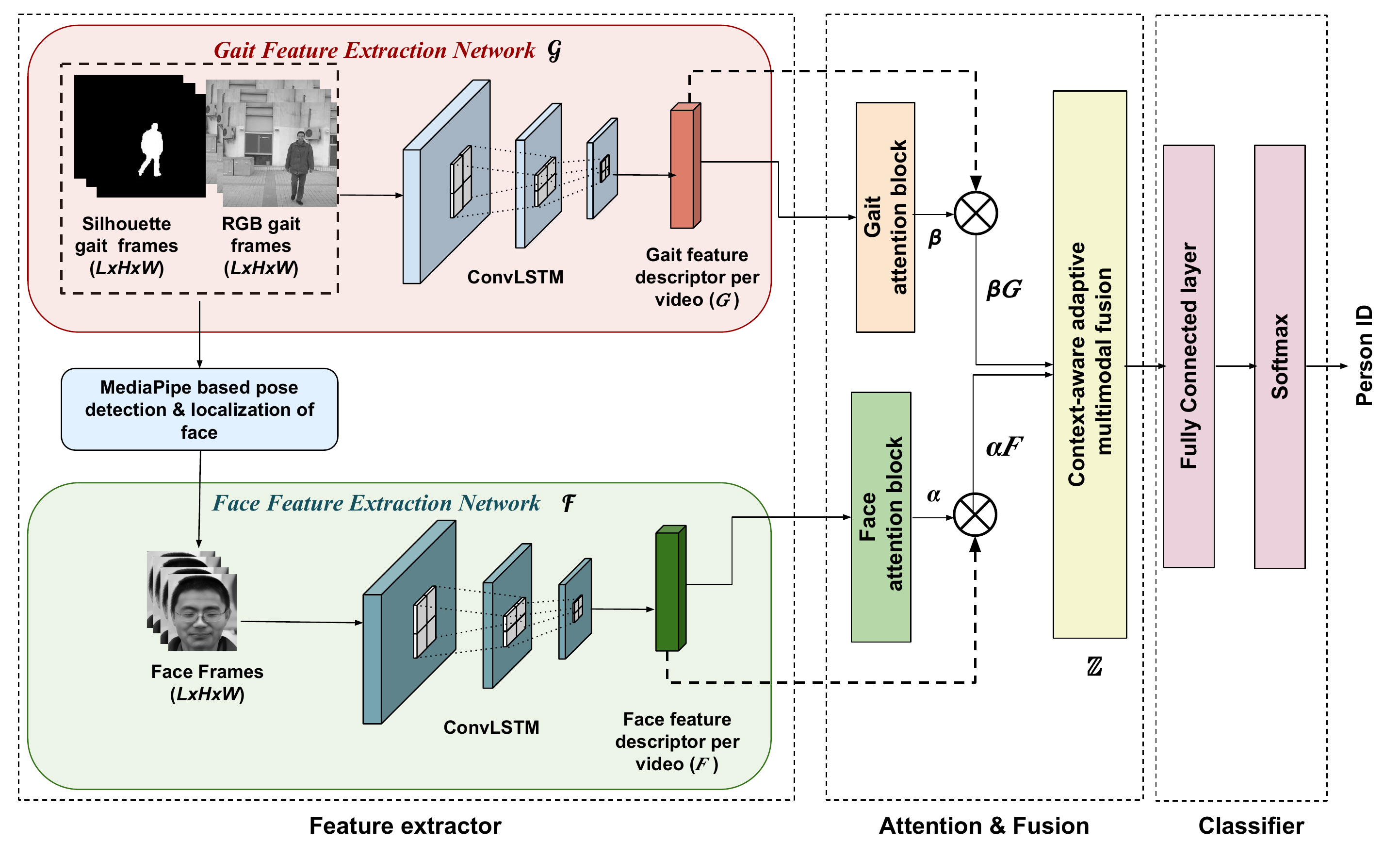}
\caption{Overall architecture of the proposed keyless attention-based adaptive fusion of face and gait for person recognition. The gait and face features are encoded by the gait and face feature extraction networks \textit{$\mathcal{F}$} and \textit{$\mathcal{G}$}, respectively. The outputs are subsequently weighted using keyless attention. Context-aware adaptive multimodal fusion is then employed to fuse global gait and facial features. Finally, the outputs are passed through the classifier to determine the class (Person ID) of the person.}
\label{fig-archi}
\end{center}
\end{figure*}

The proposed keyless attention-based adaptive fusion of face and gait towards human identification is shown in Fig.\ref{fig-archi}, in which all the symbols are introduced in the following subsections. The proposed framework maps spatio-temporal feature sequences corresponding to gait and face to a single label. First, the video sequence's descriptors of gait and face are extracted from each frame via a \textbf{\textit{Feature extractor}} module. Further, the \textit{\textbf{Attention \& Fusion}} block is employed to compute the feature importance and adaptively amalgamate them.
Finally, the class probabilities are generated by a \textbf{\textit{classifier}} module using a fully connected (FC) layer, followed by a softmax layer.




\hfill
\break


\vspace{-1cm}

\subsection{Gait feature extractor}
Gait recognition involves recognizing a person based on their gait features, i.e, movement patterns~\cite{murray1967gait}. 
The temporal variation in human silhouettes is considered by calculating the cyclic pattern of movement, commonly referred to as the \textit{gait cycle}. It can be observed that the size of the closed area between the legs and the aspect ratio in the human silhouettes are alternating periodically in a gait sequence (Refer Fig. 2(a)\& 2(b)).
Based on this notion, a complete gait cycle is determined by the number of frames between three consecutive local minima (two red points in Fig. 2(b)). The corresponding frames are extracted from the RGB images. This technique of gait cycle computation is applied to every person. Accordingly, the video is divided into an adequate number of frames required for gait feature computation.

The images are preprocessed and converted from RGB to grayscale to facilitate computational efficiency. Further, the extracted frames of gait silhouette images of height \textit{H} and width \textit{W} are fed onto a Convolutional LSTM~\cite{shi2015convolutional} architecture as depicted in gait feature extractor network \textit{$\mathcal{G}$} in Fig. \ref{fig-archi} and obtains a gait feature descriptor $G$. Formally, the gait feature sequence of a video can be represented as $G = \{g_1, \cdot \cdot \cdot , g_L\}, g_i \in \mathbb{R}^C$
where $g_i$ denotes the gait feature of frame \textit{i}, \textit{C} denotes the feature dimension, and \textit{L} denotes the number of frames.





\begin{figure}[h!]%
        \centering
        \begin{subfigure}[htbp]{0.25\textwidth}
            \centering
            \includegraphics[width=3.5cm]{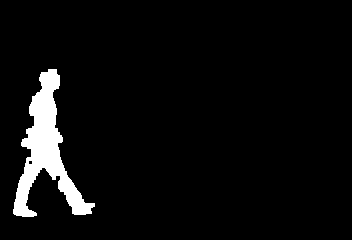}%
            \caption{}
        \end{subfigure}%
        \begin{subfigure}[htbp]{0.25\textwidth}
            \centering
            \includegraphics[width=3.5cm]{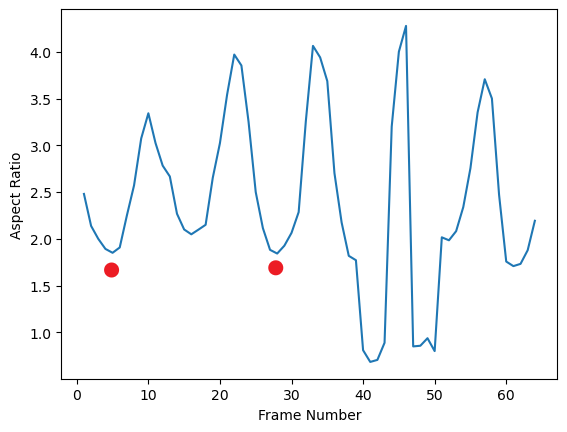}%
            \caption{}
        \end{subfigure}\break\vskip 1mm
        \begin{subfigure}[htbp]{0.165\textwidth}
            \centering
            \includegraphics[width=2.4cm]{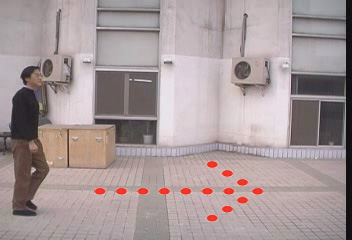}%
            \caption{0\textdegree}
        \end{subfigure}%
        \begin{subfigure}[htbp]{0.165\textwidth}
            \centering
            \includegraphics[width=2.4cm]{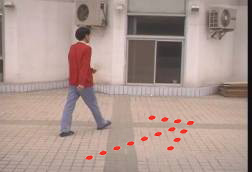}%
            \caption{45\textdegree}
        \end{subfigure}%
        \begin{subfigure}[htbp]{0.165\textwidth}
            \centering
            \includegraphics[width=2.4cm]{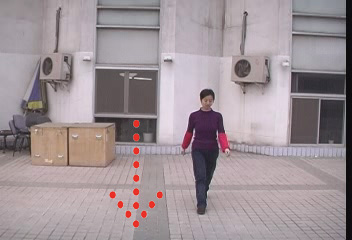}%
            \caption{90\textdegree}
        \end{subfigure}
        \caption{(a) Human silhouette taken from a gait video sequence of CASIA-A. (b) Representation of silhouette aspect ratio over the whole video. The marked points in red represent the starting and ending of one gait cycle. (c),(d) \& (e) Glimpses from the CASIA-A dataset at angles 0\textdegree, 45\textdegree, 90\textdegree \ respectively.}
        \label{gait_class}
    \end{figure}


\subsection{Face feature extractor}
Face recognition involves recognizing a person by his facial features~\cite{mm1966}. In our case, since the viewpoint and distance of the person vary significantly across the frames, traditional face detection algorithms that rely on the frontal view do not work well. Hence, facial bounding boxes are initially cropped out of the video frames leveraging Google Mediapipe human pose detection framework~\cite{bazarevsky2020blazepose}. The framework employs a two-step detector-tracker setup where the detector locates the pose region-of-interest (ROI) within the frame and the tracker predicts all 33 keypoints from this ROI. In the case of videos, the detector is run only on the first frame and the ROI of the subsequent images is derived from the pose keypoints of the previous frame.
\begin{figure}[h!]%
    \centering
    \includegraphics[trim={0 5.5cm 0 6.5cm},clip,width=9cm]{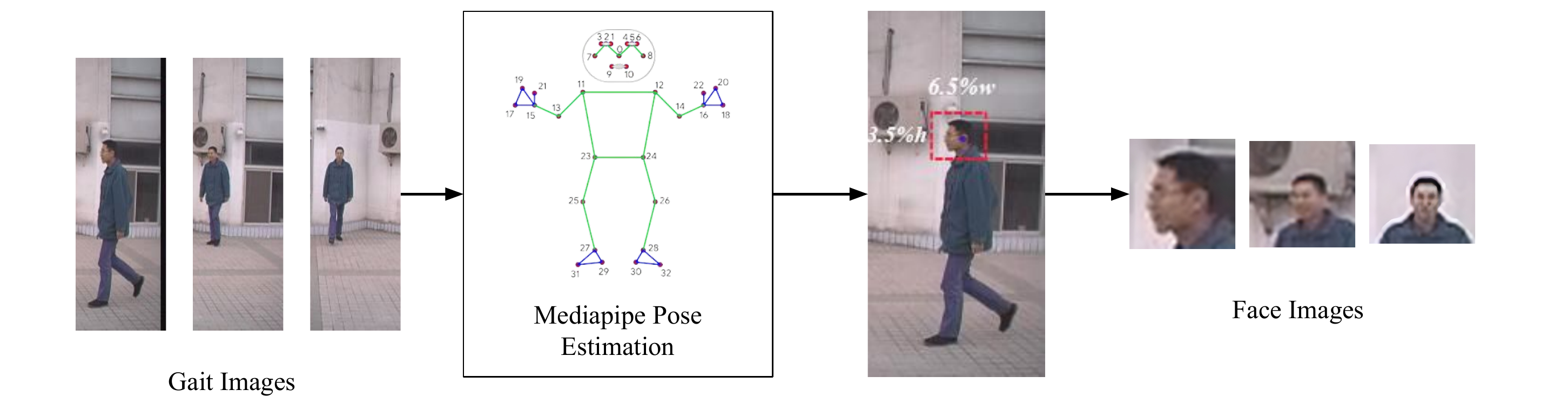}
    \caption{Process of obtaining face images from Mediapipe pose estimation. 
    }
    \vspace{-.75cm}
    \label{face_structure}
\end{figure}%
As shown in Fig.~\ref{face_structure}, from the estimated Mediapipe keypoints, the human face is manually cropped out by fixed measurements, with respect to the facial coordinates.

The cropped face images are preprocessed, converted from RGB to grayscale, and resized to the dimension of $\textit{H} \times \textit{W}$. Further, the images are fed into ConvolutionalLSTM~\cite{shi2015convolutional} architecture to extract facial feature descriptor $F$ per person, as depicted in the face feature extractor network \textit{$\mathcal{F}$} in Fig.\ref{fig-archi}. Formally, the face feature descriptor corresponding to an \textit{L}-frame video is represented as $F = \{f_1, \cdot \cdot \cdot, f_L\}, f_i \in \mathbb{R}^C$, where $f_i$ represents the facial feature of frame \textit{i}. 


\subsection{Na\"ive fusion of Face and Gait via Bilinear pooling}\label{naive-fusion}

As an initial fusion technique, we propose the na\"ive bilinear pooling (BLP) method~\cite{lin2015bilinear} to fuse features. The method takes in the 3D tensor outputs from the final max-pooling layers of the face (\textit{$\mathcal{F}$}) and gait  (\textit{$\mathcal{G}$}) feature extraction networks. The outputs are further reshaped into the matrix of dimensions $p\times d$ and are combined  via the bilinear pooling method to obtain the fusion result $Z$, as follows:
\break
\vspace{-0.75cm}

\begin{equation}
    Z = FG^{T}, F \in \mathbb{R}^{p\times d}, G \in \mathbb{R}^{p\times d}
\end{equation}

The matrix $Z$ is then flattened into a vector and then passed onto the softmax activation function, where it computes the probability for class \(k\)  out of \(K\) classes.

\subsection{Keyless attention based Adaptive fusion of Face and Gait}
\label{attention}

Attention mechanisms are widely used in sequence models, enabling the modelling of dependencies regardless of their location in the input or output sequences~\cite {bahdanau2014neural}. In our case, not every frame in a video helps identify a subject in the same way. In order to estimate the importance weights for each frame, we adapt the attention mechanism. An attention function is a process that takes a query vector and a set of key-value pairs and produces an output.
In existing soft attention mechanisms~\cite{bahdanau2014neural}, the weight computation is not limited to the feature vectors but also incorporates an additional input, such as the previously hidden state vector of the LSTM or a vector representing a target entity as in~\cite{wang2016relation}. These additional inputs along with feature vectors referred to as \textit{key vectors}, help to 
find the most related weighted average of feature vectors. However, the weights in our work depend only on the feature vectors and do not require any additional input, thus named as \textbf{\textit{keyless attention}}, synonymous to the work \cite{long2018multimodal}. 
In our case, referring to Fig. \ref{fig-archi}, the gait feature descriptor \textit{G} and face feature descriptor \textit{F} are further fed onto two attention modules \textit{viz.} Gait attention block and Face attention block respectively. Further, multimodal adaptive weights are computed via the fusion mechanism. Detailed explanations are given below.



\subsubsection{\textbf{Face Attention}}\label{fatt}

The facial feature is updated by incorporating the attention mechanism to assign weighted visual elements. Formally, face attention is computed as follows:
\vspace{-0.25cm}
\begin{equation}
\Bar{f_i} = \mathbf{W_f} f_i + \mathbf{b_f}
\end{equation}
\vskip -4mm
\begin{equation}
\Bar{e_i} = \Bar{u}^T \tanh(\Bar{f_i})
\end{equation}
\vskip -3mm
\begin{equation}
\Bar{\alpha_i} = \frac{\exp{(\lambda \Bar{e_i})}}{\sum_{k=1}^{L} \exp{(\lambda \Bar{e_k})}}
\end{equation}
Here, $\Bar{f_i}$ is the low-dimension representation of frame \textit{i} and $\mathbf{W_f}$ \& $\mathbf{b_f}$ are the learnable parameters. The importance weight, $\bar{e_i}$, of the element $f_i$ is computed by the inner product between the new representation of $\bar{f_i}$ and a learnable vector $\bar{u}$. The normalized importance weight of facial feature $\Bar{\alpha_i}$ is calculated using the softmax function, as shown in Eq. (5).  $\lambda$ is a scale factor that ensures that the importance weights are evenly distributed, which ranges between 0 and 1. Nevertheless, it is observed in Eq.~(4) that $\tanh{(\cdot)}$ non-linearity may not be effective for learning complicated linkages, since $\tanh{(x)}$ is roughly linear for x $\in $ [$-1$, 1]. Therefore, inspired by the method as in~\cite{DauphinFAG16}, we leverage an effective gated mechanism as shown in Eq.~(6) to formulate a better normalized facial importance weight $\Bar{\alpha_i}$.
\vspace{0mm}
\begin{equation}
\Bar{\alpha_i} = \frac{\exp{\{\lambda \Bar{u}^T (\tanh(\Bar{f_i}) \odot sigm(\Bar{f_i})\}}}{\sum_{k=1}^{L} \exp{\{\lambda \Bar{u}^T (\tanh(\Bar{f_k}) \odot sigm(\Bar{f_k})\}}}
\end{equation}
\vspace{-2mm}
\begin{equation}
\Bar{\alpha} = \sum_{i=1}^{L} \Bar{\alpha_i}
\label{alpha}
\end{equation}%

where \textit{sigm($\cdot$)} is the sigmoid non-linearity and $\odot$ is an element-wise multiplication. This new $\Bar{\alpha_i}$ is further used to compute global facial attention weight 
$\Bar{\alpha}$ by combining facial importance weight across all \textit{L} frames(Refer Eq.~(7)).  


\vskip -3mm


\subsubsection{\textbf{Gait Attention}}\label{gatt}
Analogous to the face modality, the attention mechanism is incorporated in the gait counterpart as well. The global gait attention weight $\Bar{\beta}$  by leveraging the weighted visual elements in the gait stream is computed as follows.
\vspace{-0.15cm}
\begin{equation}
\Bar{g_i} = \mathbf{W_g} g_i + \mathbf{b_g}
\end{equation}
\begin{equation}
\Bar{\beta_i} = \frac{\exp{\{\lambda \Bar{u}^T (\tanh(\Bar{g_i}) \odot sigm(\Bar{g_i})\}}}{\sum_{k=1}^{L} \exp{\{\lambda \Bar{u}^T (\tanh(\Bar{g_k}) \odot sigm(\Bar{g_k})\}}}
\end{equation}
\vskip -2mm
\begin{equation}
\Bar{\beta} = \sum_{i=1}^{L} \Bar{\beta_i}
\label{beta}
\end{equation}

\subsubsection{\textbf{Context-aware Adaptive Fusion}}\label{adap-fusion}
From Eq.(\ref{alpha}) \& Eq.((\ref{beta}), we obtain the value of $\Bar{\alpha}$  and $\Bar{\beta}$, which are the global individual attention weights of the face and gait features, respectively. The weighted average of the face and gait feature is computed using the adaptive weights:\vspace{-2mm}
\begin{equation}
    \alpha = \frac{\norm{\Bar{\alpha}}}{\norm{\Bar{\alpha}} + \norm{\Bar{\beta}}}
\end{equation}
\begin{equation}
    \beta = \frac{\norm{\Bar{\beta}}}{\norm{\Bar{\alpha}} + \norm{\Bar{\beta}}}
\end{equation}
The adaptive fusion is performed by combining the two features multiplied individually by their weighted global attention weights, as follows:
\vspace{-0.4cm}
\begin{equation}
    \textit{$\mathbb{Z}$} = \alpha F + \beta G
\end{equation}
\vskip -2mm
where \noindent{ $\mathbb{Z}$ refers to the context-aware adaptively fused feature. }
$\mathbb{Z}$ is further passed onto a fully-connected (FC) layer, followed by a softmax function that classifies the feature according to the \textit{K} classes provided. The resultant column vector, $R$ is then used to determine the class identifier (\textit{Person ID}) of the subject in consideration for the fused feature $\mathbb{Z}$ by\\ \vspace{-0.5cm}
\begin{equation}
    ID(\mathbb{Z}) = argmax(R)
\end{equation}
\vspace{-1cm}
\subsection{Objective functions}\label{obj-func}
The model classifier employs categorical cross-entropy loss, also known as Softmax loss. This supervised loss calculates the classification error among \textit{K} classes. The number of nodes in the softmax layer depends on the number of identities in the training set. Considering \(t\) and $w_k$ as the target vector and learnable vector respectively, the loss is computed as:
\vspace{-0.1cm}
\begin{equation}
    Loss = -\sum_{i}^{K} t_i\log(softmax(w_k^T\mathbb{Z})_i)
\end{equation}


\section{Experimental Setup}
\noindent \textbf{Dataset:} In this work, we use CASIA Gait Dataset A~\cite{wang2003silhouette}, which includes 19139 images of 20 subjects. Each person has 12 image sequences, 4 sequences for each of the three directions, i.e. 
0\textdegree, 45\textdegree, 90\textdegree (Refer Fig. 2(c), 2(d) and 2(e)). Among the 4 sequences per angle, 2 sequences are used for training, and the remaining 2 sequences are used for testing.

\noindent \textbf{Evaluation protocols:} Standard evaluation metrics like \emph{accuracy} and \emph{log-loss} are employed to validate the performance of our model. \emph{Accuracy} is used to evaluate how well the algorithm is performing for all classes by giving them equal importance, whereas \emph{log loss} is considered to be a crucial metric that is based on probabilities. Mathematically, log-loss is computed by:
\begin{equation}
log\_loss = \frac{-1}{N \sum_{i=1}^{N} [y_i\ln p_i+(1-y_i)\ln(1-p_i)]}
\end{equation}
where \textit{N} is number of person and $y_i$ is the observed value and $p_i$ is predicted probability.

\noindent \textbf{Implementation details:} The proposed method is implemented using the TensorFlow framework.
During training, video frames correspond to one gait cycle across three orientations 0\textdegree, 45\textdegree, and 90\textdegree are considered. In this work, the gait cycle corresponds to
\textit{L} = 24 frames, each with height $\textit{H} $= 128, and width $\textit{W}$ = 128 is used. The images are normalized using the RGB mean and standard deviation of ImageNet before passing them to the network. After dimension reduction, the resulting dimension of the gait and face feature descriptor is 588 each. In the experiments, we use Optuna~\cite{akiba2019optuna}, a hyperparameter optimization framework, to obtain the best hyperparameter for our models. 
We train the network for approximately 1000 iterations. The implementations are done in a machine with Tesla V100 GPU with 12GB RAM and took around 1 hour to train the model.
\section{Experimental Results}\vspace{-0.1cm}
To verify the effectiveness of our proposed approach, various experiments using single feature-based and multimodal fusion-based human identification are carried out. The result summary is shown in Tables I and II. Referring to Table I, the first two rows are single-modality based results, whereas the remaining are multi-modality results.\\
\noindent \textbf{(i) Face feature based human recognition:}
Training of facial features separately on each angle 0\textdegree, 45\textdegree and 90\textdegree ~with custom parameters and hyperparameter tuning using Optuna~\cite{akiba2019optuna} produce accuracies up to 65\%, 80\%, and 85\%, respectively (Refer Table II). The overall accuracy of the face model incorporating all orientations is 80\% (Refer Table I).\\
\textbf{(ii) Gait feature based human recognition:} 
Training of the gait features across three view-points 
produces accuracies 75\%, 60\%, and 55\%, respectively (Refer Table II). Referring to Table I, the overall accuracy of the gait model across viewpoints is observed to be 70\%.\\
One noteworthy observation from the aforesaid single-modality based results, is the outperformance of face and gait models in 90\textdegree and in 0\textdegree viewpoints, respectively. This accentuates our initial intuition of the influence viewpoint on feature modalities. To incorporate the best of both modalities in different viewpoints, we facilitate fusion techniques. In particular, four fusion approaches are carried out.\\
\textbf{(iii) Average based fusion:}
Average weight-based fusion incorporates a manual weight input to the face and gait model.
For this technique, a weightage of 0.5 is devised on the individual face and gait features, achieving an accuracy of $75\%$ on the test dataset. 
\\
\textbf{(iv) Na\"ive fusion via BLP:}
Bilinear Pooling incorporates the fusion of  both gait and face models, as explained in Section \ref{METHOD} (C). The model achieves 85\%, 75\%, and 85\% viewpoint-wise accuracies, as shown in Table II. The overall fused model results in an accuracy of 80\%. It was observed that compared to the \textit{Average based fusion} model, \textit{Na\"ive fusion with BLP} improves the accuracy by $5\%$.\\
\textbf{(v) Attention Fusion:}
In this model, the keyless attention mechanism (discussed in Sec \ref{attention}) is implemented to obtain global face and gait attention weights \textit{$\Bar{\alpha}$} and \textit{$\Bar{\beta}$}. It is 
further multiplied with the respective features \textit{F} and \textit{G} and is concatenated to obtain a single feature vector. Note that no adaptive fusion strategy is employed in this scheme. This model is able to achieve an overall accuracy of $85\%$ incorporating features over all the viewpoints.
\\
\textbf{(vi): Context-aware Adaptive Fusion with attention:}
This strategy incorporates the proposed context-aware adaptive fusion strategy to the attention module, discussed in Sec.\ref{attention}. The viewpoint-wise accuracy attained by this method are 90\%, 80\%, and 90\%, respectively. Referring to Table I, this model outperforms all other models by achieving an overall accuracy of 90\%, highlighting the importance of the context-aware fusion of modalities across the view-points. In terms of the log loss metric, the adaptive fusion strategy has achieved the least value with 0.389 compared to all other models.





\vskip -3mm

\begin{center}
\centering{\textsc{TABLE I:} Overall result summary of the models}
\end{center}
\newcommand*\pct{\scalebox{.9}{\%}}
\begin{tabular}{l@{\hskip -1.5mm}c@{\hskip -2mm}c@{\hskip 2mm}c}\toprule
\small
    \textbf{Index} & \textbf{Model} & \textbf{Accuracy\((\pct)\)}  & \textbf{Log Loss}\\ \midrule
    (i)&Face feature model & 80 & 0.436\\
    (ii)&Gait feature model & 70 & 1.641\\
     \midrule
     (iii)&Average based fusion  &  75 & 0.779\\
     (iv)&Na\"ive Fusion via BLP  &  80 & 0.519\\
     (v)&Attention Fusion  &  85 & 1.619\\
     (vi)&\textbf{Adaptive Fusion Attention}  &  \textbf{90} & \textbf{0.389}\\
     \midrule
     (vii)&\textbf{\shortstack{Geng, Wang et. al.~\cite{geng2008adaptive}}} & 86.67 & -
     \\\bottomrule
\end{tabular}

In order to demonstrate the effectiveness of the proposed \textit{adaptive fusion of gait and face with attention} algorithm, a comparative analysis against the state-of-the-art result on CASIA-A dataset is carried out. The experimental result in Table I (vii) shows that our adaptive fusion result (90\%) outperforms the state-of-the-art performance reported by Geng et al.~\cite{geng2008adaptive}, which had an overall average test accuracy of 86.67\%. Our attention fusion model is also found to be achieving competitive result (85\%) with the state-of-the-art result. These results clearly highlight the potential of our proposed attention-based adaptive multimodal fusion using deep neural networks, in contrary to their
classical hand-crafted features and manual condition-based fusion approach.

\vspace{0.3cm}
\begin{center}
\centering{\textsc{TABLE II:} Result of all models conducted angle-wise $0^{\circ}$, $45^{\circ}$, and $90^{\circ}$ with respect to camera}
\end{center}
 \vspace{-0.3cm}   
\begin{table}[!h]
\centering
\footnotesize
\renewcommand{\arraystretch}{1.2}
\setlength\tabcolsep{5pt}
 \begin{tabular}{|c|c|c|c|c|}
 \hline
\multirow{2}{*}{\textbf{Angle}($^{\circ}$)} & \multicolumn{4}{c|}{\textbf{Accuracy\((\%)\)}} \\\cline{2-5}
& \textbf{Face} & \textbf{Gait} & \textbf{Na\"ive Fusion} & \textbf{Adaptive Fusion}\\\hline
0\textdegree & 65 & 75 & 85 & 90 \\\hline
45\textdegree & 75 & 60 & 75 & 80 \\\hline
90\textdegree & 85 & 55 & 85 & 90 \\\hline
\hline
\end{tabular}
\label{tab:table2}
\end{table}

From the view-point-wise performance of the models in Table II, some key interpretations also can be drawn. For 0\textdegree, the gait model surpasses the face model performance, which aligns with our intuition that the model learns gait features better when the subject walks laterally. Similarly, works well when the subject walks towards the camera at 90\textdegree. However, while incorporating adaptive fusion strategy, the best of both modalities are incorporated adaptively based on the context, resulting in high performance irrespective of the view-point.


\begin{figure}[htbp]
    \centering
   \includegraphics[width=0.27\textwidth]{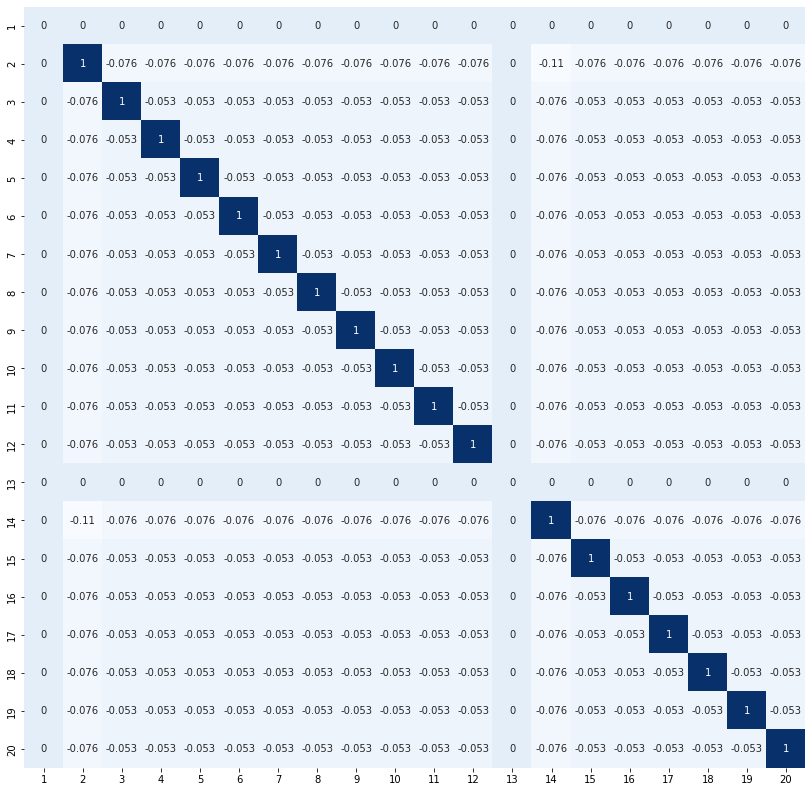}
    \caption{Confusion matrix of adaptive fusion
attention model on the test set}
    \label{corr_matrix}
\end{figure}   

A visual interpretation of the attention based adaptive fusion model is depicted in Fig. \ref{corr_matrix} in terms of a confusion matrix. A confusion matrix visualizes and summarizes the performance of a classification algorithm. All the three viewpoints together on the test dataset of 20 subjects is depicted in Fig. \ref{corr_matrix}. We can observe that 18 out of 20 subjects are correctly classified, except subject IDs `1' and `13'. The subject ID `1' is most often classified as 14 and the subject ID `13'  as 5 and 20 in most models. Further, we observe that subject 13 being identified correctly by the face model but not in the gait model, which might be  ascribable to the suboptimal learning of the model from the grayscale features of the image.




\section{Conclusion and Future Works}
In this work, we proposed a multimodal adaptive fusion of the face and gait toward human identification. In particular,  a keyless attention-based deep neural network for learning the attention in the  gait and face videos and a context-aware adaptive fusion strategy to efficiently extract and fuse the features is presented. Based on the observation that single biometric modality results in suboptimal results, various studies leveraging average based, na\"ive fusion, attention fusion and context-aware adaptive fusion were investigated. Results of the proposed attention-based adaptive fusion strategy show superior performance compared to all the other models as well as the state-of-the-art result. Future Improvements can be made by introducing better attention mechanisms such as dense co-attention and spatial-channel attention, as well as advanced fusion mechanisms like tucker fusion, block fusion etc.
\printbibliography
\end{document}